\documentclass[sigconf]{acmart}
\AtBeginDocument{%
  }

\copyrightyear{2025}
\acmYear{2025}
\setcopyright{acmlicensed}
\acmConference[MM '25] {Proceedings of the 33rd ACM International Conference on Multimedia}{October 27--31, 2025}{Dublin, Ireland.}
\acmBooktitle{Proceedings of the 33rd ACM International Conference on Multimedia (MM '25), October 27--31, 2025, Dublin, Ireland}
\acmISBN{979-8-4007-2035-2/2025/10}
\acmDOI{10.1145/3746027.3754848}

\usepackage{tabularx}
\usepackage{multirow}
\usepackage{algorithm} 
\usepackage{algpseudocode}

\begin{document}

\title{OnlineHOI: Towards Online Human-Object Interaction Generation and Perception}


\author{Yihong Ji}
\email{jiyihong2021@email.szu.edu.cn}
\orcid{0009-0004-5033-7530}
\affiliation{%
  \institution{College of Computer Science and Software Engineering, \\ Shenzhen University, \\ Guangdong Laboratory of Artificial Intelligence and Digital Economy (SZ)}
  \country{China}
}

\author{Yunze Liu}
\email{liuyzchina@gmail.com}
\orcid{0009-0002-3148-8822}
\affiliation{%
  \institution{Tsinghua University, \\
  Shanghai Qi Zhi Institute}
  \country{China}
  }

\author{Yiyao Zhuo}
\email{zhuoyiyao@gml.ac.cn}
\orcid{0009-0004-7120-0808}
\affiliation{%
  \institution{Guangdong Laboratory of Artificial Intelligence and Digital Economy\\
  (SZ)}
  \country{China}
}

\author{Weijiang Yu}
\email{weijiangyu8@gmail.com}
\orcid{0000-0002-7449-3093}
\affiliation{%
 \institution{Sun Yat-Sen University}
  \country{China}
 }

\author{Fei Ma}
\authornote{Corresponding Author.}
\orcid{0009-0002-5388-9125}
\email{mafei@gml.ac.cn}
\affiliation{%
  \institution{Guangdong Laboratory of Artificial Intelligence and Digital Economy (SZ)}
  \country{China}
  }

\author{Joshua Zhexue Huang}
\email{zx.huang@szu.edu.cn}
\orcid{0000-0002-6797-2571}
\affiliation{%
  \institution{Shenzhen University}
  \country{China}
  }

\author{Fei Yu}
\email{yufei@szu.edu.cn}
\orcid{0000-0003-1006-7594}
\affiliation{%
  \institution{Guangdong Laboratory of Artificial Intelligence and Digital Economy (SZ)}
  \country{China}
  }

\renewcommand{\shortauthors}{Yihong Ji et al.}

\begin{abstract}
The perception and generation of Human-Object Interaction (HOI) are crucial for fields such as robotics, AR/VR, and human behavior understanding. However, current approaches model this task in an offline setting, where information at each time step can be drawn from the entire interaction sequence. In contrast, in real-world scenarios, the information available at each time step comes only from the current moment and historical data, i.e., an online setting. We find that offline methods perform poorly in an online context. Based on this observation, we propose two new tasks: Online HOI Generation and Perception. To address this task, we introduce the OnlineHOI framework, a network architecture based on the Mamba framework that employs a memory mechanism. By leveraging Mamba's powerful modeling capabilities for streaming data and the Memory mechanism's efficient integration of historical information, we achieve state-of-the-art results on the Core4D and OAKINK2 online generation tasks, as well as the online HOI4D perception task.
\end{abstract}


\begin{teaserfigure}
\centering
\includegraphics[width=0.8\linewidth]{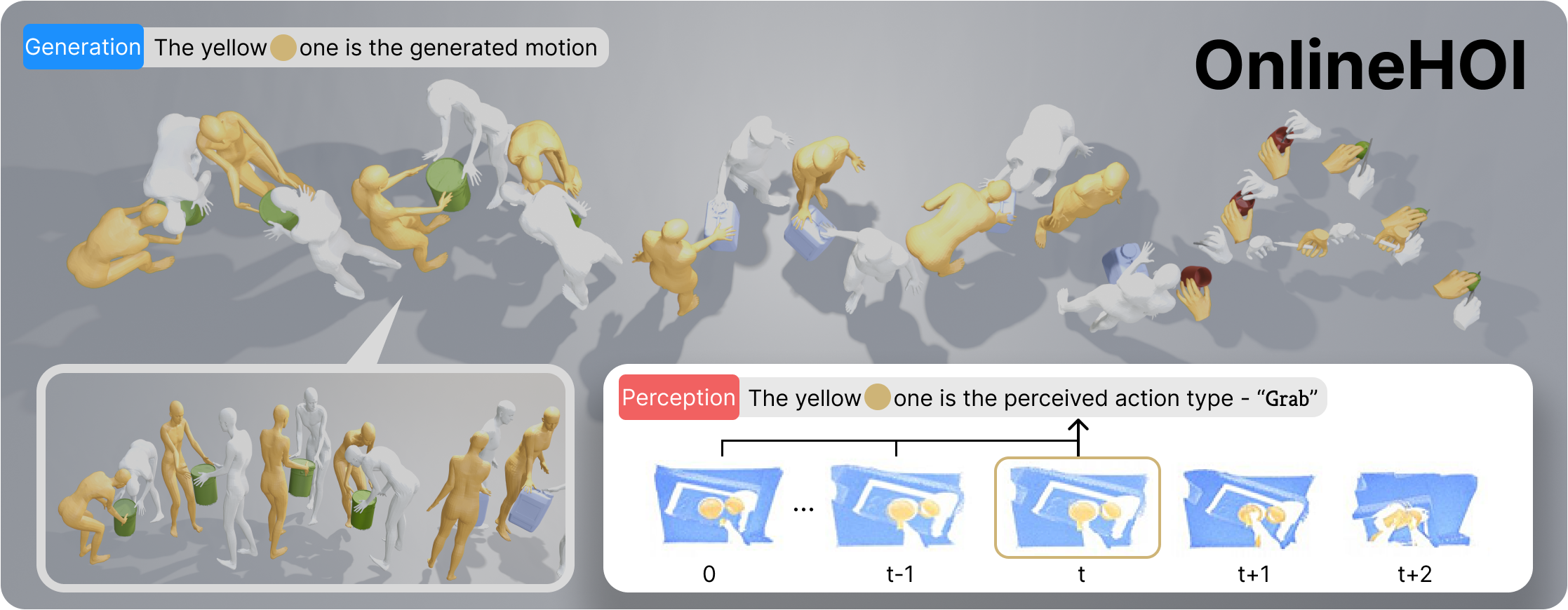}
\label{fig:begin}
\caption{We propose OnlineHOI for the online generation and perception of human-object interactions (HOI). Unlike traditional offline settings, the online setting requires that the network only access information from historical frames and cannot perceive future frames, posing new challenges for existing models. We introduce an OnlineHOI framework that can significantly improve the performance of online HOI generation and perception. In the online generation task, the yellow-highlighted person or hand is the reactor, whose movements are generated based on the actions of another person or hand. The online perception task utilizes the current streaming input to predict the category of hand motion interactions at the current time step.}
\end{teaserfigure}


\maketitle

\section{Introduction}
\label{sec:intro}

Human-Object Interaction (HOI) perception and generation are essential for developing intelligent systems. They enable robots and AR/VR applications to accurately understand human actions and object usage in real-time, improving collaboration and safety. Additionally, HOI generation supports realistic simulation and prediction of interactions, facilitating training and immersive experiences. Together, these technologies enhance system performance, foster innovative research, and bridge the gap between the digital and physical worlds, ultimately leading to more adaptive and human-centric designs in various fields \cite{xue2024human,xie2025pointtalk,feng2025unisync,ma2025review} for future progress.

Current methods for Human-Object Interaction (HOI) perception \cite{fan2021point,wen2022point} and generation \cite{li2023object,tevet2022human,li2024controllable} predominantly rely on an offline modeling paradigm, wherein complete interaction sequences are available for processing. However, real-world scenarios often require an online approach, as only historical and current information is accessible during dynamic interactions, such as collaborative tasks \cite{liu2024core4d}. This discrepancy introduces significant challenges: online settings must handle incomplete data streams, increased uncertainty due to the absence of future context, and the necessity for real-time processing. Consequently, there is a pressing need to develop robust, adaptive models that integrate memory mechanisms to infer and predict interactions accurately in a sequential, online environment.

\begin{figure}
\centering
\includegraphics[width=0.35\textwidth]{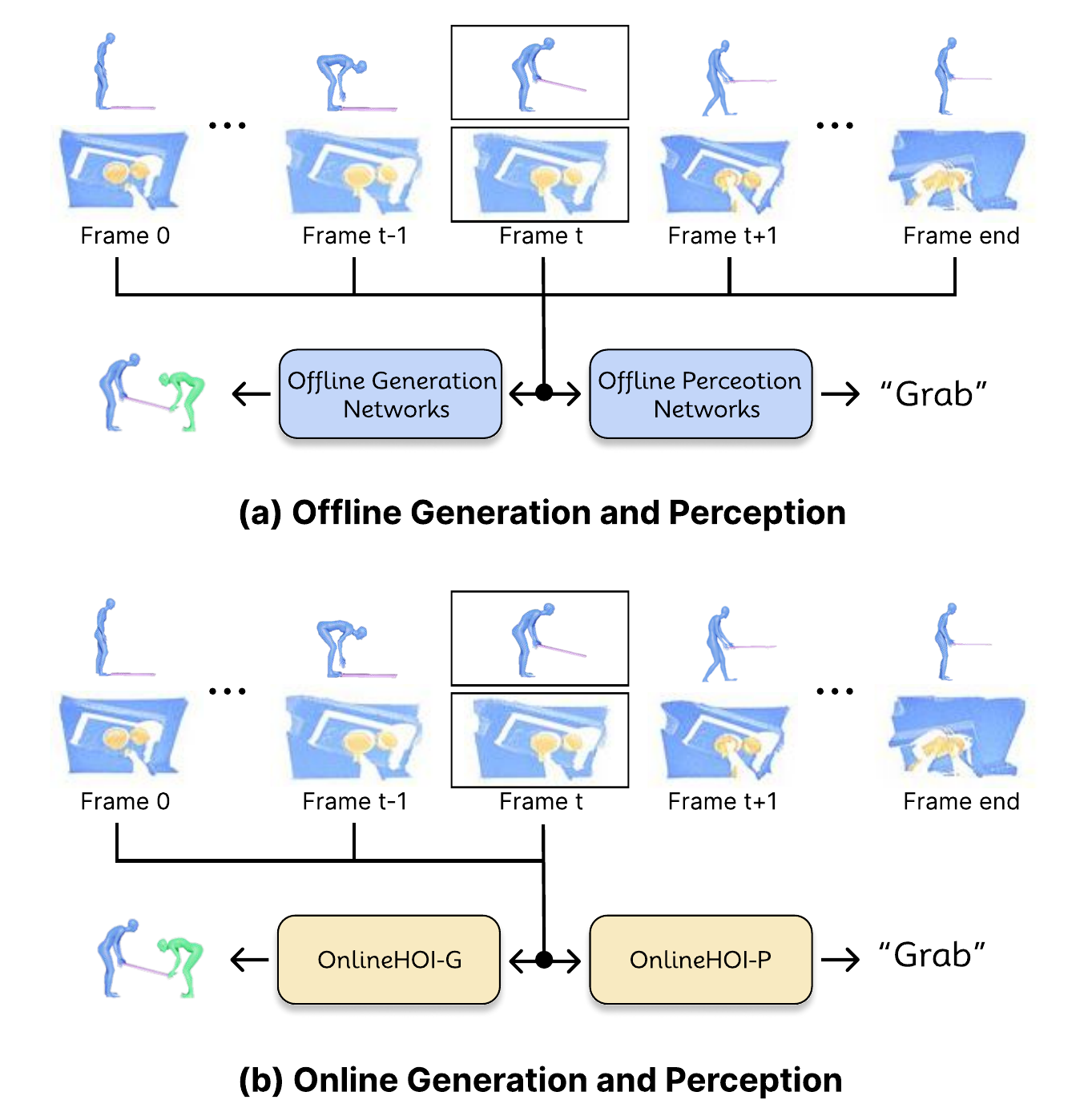}
\caption{Comparison of the Offline and Online settings. (a) The whole timeline is known. (b) The network can only make predictions based on the current frame and previous information.}
\vspace{-7mm}
\label{fig:offline and online}
\end{figure}

Our core observation is that Transformer-based \cite{vaswani2017attention} architectures perform poorly in online settings. We believe this is due to their limited ability to model variable-length data and their insufficient capacity to capture information from consecutive frames. In contrast, the Mamba \cite{gu2023mamba} architecture, with its continuous scanning mechanism and superior long-sequence modeling capabilities, is better suited for online tasks. Moreover, to further enhance the integration of historical information, we propose a Memory mechanism to boost network performance.

We propose a Mamba Memory Augment Model named OnlineHOI for online generation and perception. The OnlineHOI incorporates two key components as shown in Figure. \ref{fig:overview}: (1) Mamba Model: This model is ingeniously crafted to organize motion frames in a sequential manner, leveraging unidirectional spatial scanning mechanisms on an online setting. These scanning mechanisms are designed to optimize unidirectional information propagation, enabling seamless top-down processing. (2) Memory Augment Model: This model is designed to enhance the historical Memory, which consists of short-term memory and long-term memory. These two distinct types of knowledge are then integrated through concatenation to encode the memory information. The enhanced memory module improves the Mamba model's ability to predict future states in online settings.

We conduct experiments on Core4D \cite{liu2024core4d}, OKINK2 \cite{yang2022oakink} and HOI4D \cite{liu2022hoi4d} dataset. In the generation task, we implement OnlineHOI-G on Core4D and OKINK2 dataset and achieved performance improvements in terms of Fréchet Inception Distance (FID), Diversity (DIV). In the perception task, we execute OnlineHOI-P and realized a significant performance boost of Acc, Edit, and F1 scores. Moreover, comprehensive ablation studies validated the effectiveness of our approach.

Our contributions are three-fold. 
\textbf{First}, we introduce two new tasks: online HOI perception and generation. These tasks pose new challenges for feature extraction from streaming data and impose new requirements on network architectures.
\textbf{Second}, we propose OnlineHOI-G and OnlineHOI-P to enhance the performance of online HOI perception and generation tasks. Our approach features a Mamba-based network architecture coupled with multi-range Memory to bolster the extraction of historical information, thereby enabling effective feature extraction from streaming data. 
\textbf{Third}, we demonstrate the effectiveness of our method through extensive experiments on both generation and perception tasks, achieving state-of-the-art results on the Core4D, OAKINK2, and HOI4D datasets. Comprehensive ablation studies further validate the rationality of our design.

\section{Related Work}
\label{sec:related}
\subsection{Human Object Interaction Generation}
Generating human motion has garnered significant attention in 3D animation, particularly in works focused on the domain of text-to-motion \cite{guo2022tm2t, petrovich2022temos, tevet2022human, tevet2022motionclip, zhou2023ude, zhang2023generating, chen2023executing, zhang2024motion}. Additionally, some studies address the problem of human-scene interaction \cite{wang2022humanise, zhao2023synthesizing, xuan2023narrator, huang2023diffusion, jiang2024scaling} and human-object interaction \cite{cha2024text2hoi, li2023object, he2024syncdiff, dai2024interfusion, li2024controllable}. Human-object interaction generation is a conditional task that generates human motion based on a dynamic object. Early work, such as Couch \cite{zhang2022couch}, employed conditional variational autoencoders to model the distribution of human-object interactions. With the remarkable advancements in diffusion models \cite{ho2020denoising}, diffusion-based methods \cite{peng2023hoi, diller2024cg, li2024controllable} have been widely proposed to handle more complex interactions. MDM \cite{tevet2022human} introduced a transformer-based conditional diffusion model to generate full-body motion. OMOMO \cite{li2023object} proposed a two-stage method consisting of two transformer-based conditional diffusion frameworks: the first generates contact information from an object motion sequence, while the second generates human motion. More recently, several studies have focused on more complex tasks, such as multi-human or multi-object interaction generation \cite{lv2024himo, daiya2024collage, he2024syncdiff}. Multi-person and object collaboration synthesis, such as \cite{daiya2024collage, he2024syncdiff}, generate multi-human HOI using LLM guidance or frequency domain decomposition. Our method focuses on more sophisticated conditional tasks, such as generating human-object interactions where, in addition to the human interacting with the object, another human must be reasonably generated to handle tasks like collaboration or handover. While the aforementioned methods primarily address the problem in an offline setting, they may not adapt well to the online setting. Therefore, we propose a Mamba-based backbone with a memory design that outperforms previous offline methods.

\subsection{Human Object Interaction Perception}
Point cloud video understanding is crucial for modeling the 4D world, combining 3D spatial and 1D temporal data. Understanding actions and interaction prediction from 2D videos or images can also be explored within multimodal Large Language Models (LLMs) that enhance integrated vision-language reasoning in a unified framework. P4Transformer \cite{fan2021point} utilizes spatiotemporal convolutions to extract local features from the entire sequence, followed by Transformers for broader spatiotemporal information exchange. PPTr \cite{wen2022point} addresses the efficiency limitations of P4Transformer by introducing primitive planes as a compact mid-level representation, improving computational efficiency while preserving essential structural information. Recently, vision-language models (VLMs) have been leveraged to handle this task. HANDSONVLM \cite{bao2024handsonvlm} predicts low-level actions for the future by initially inferring prospective high-level actions using a vision-language model (VLM). However, while these 4D networks and VLMs are generally designed for offline settings, they struggle to handle online settings effectively due to the limitations of Transformers when modeling variable-length, unidirectional sequences. This limitation motivated us to explore a new architecture based on the Mamba model, integrating a Memory Augment mechanism that is well-suited for online perception. The results demonstrate that our method, designed specifically for the online setting, significantly improves performance compared to offline methods. We believe that our approach can also be beneficial for adapting VLMs to various downstream applications in online settings.

\subsection{Memory-augmented Understanding}
The memory model has been applied in various tasks, such as long video understanding \cite{song2024moviechat}, video object segmentation \cite{cheng2022xmem, seong2021hierarchical}, visual object tracking \cite{ma2018adaptive, zhou2023memory}, and action understanding \cite{wang2023memory}. MovieChat \cite{song2024moviechat} introduces a memory mechanism that transfers information over a large temporal range to reduce redundancy in visual tokens and enhance long-term dependencies. XMem \cite{cheng2022xmem} proposes an architecture that utilizes multiple independent yet highly interconnected feature memory modules to efficiently process long videos containing thousands of frames. MeMOT \cite{cai2022memot} constructs a large spatiotemporal memory to retain past observations of tracked objects. Since our focus is on online HOI, we first validate that the Mamba-based model outperforms the Transformer-based model in the online setting. However, the Mamba model faces limitations in retrieval and indexing, which are critical for learning from long online sequences. To address this, we propose a Memory-Augmented system that enhances online understanding. Our results show that this system significantly improves the Mamba model's ability to handle tasks in the online setting.

\section{Method}
\subsection{Overview}
\begin{figure*}
\centering
\includegraphics[width=0.81\linewidth]{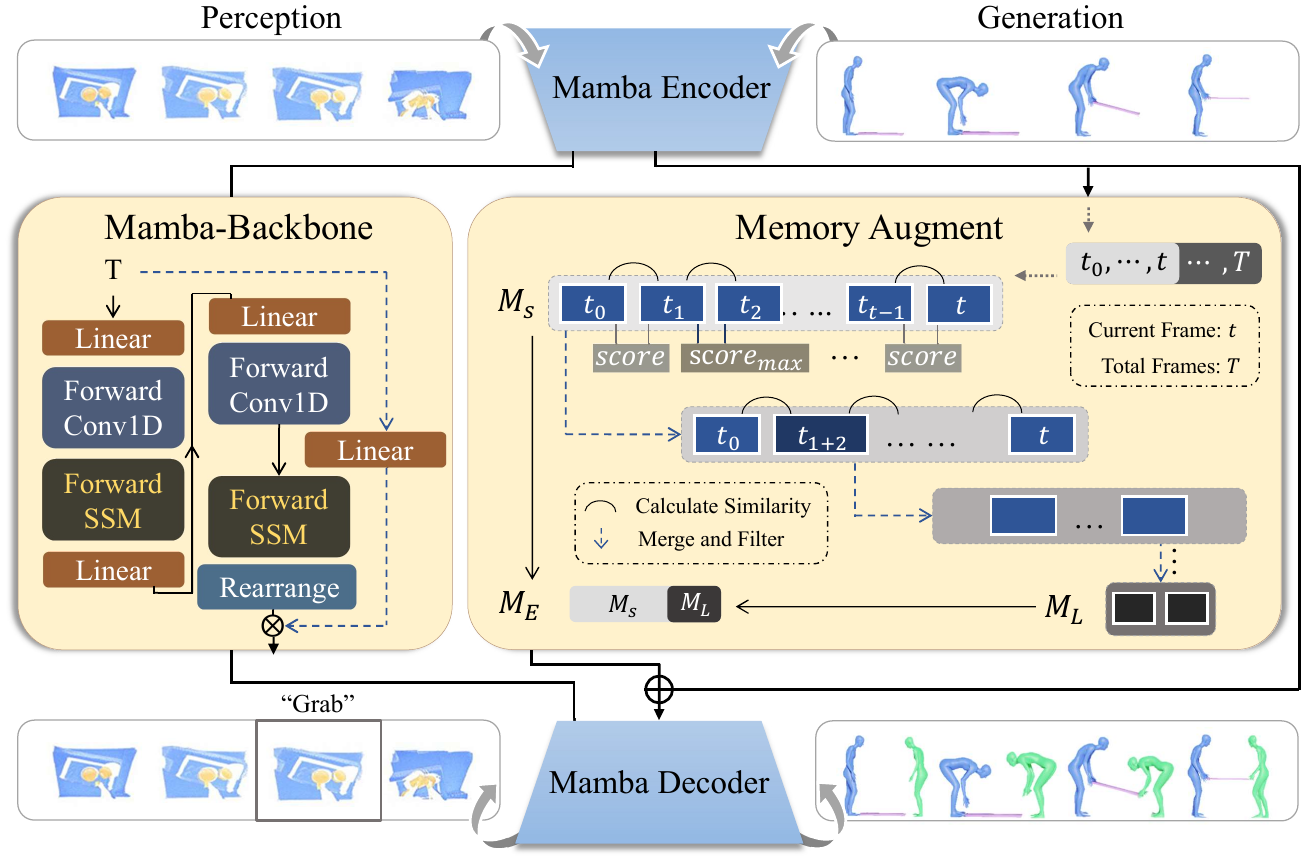}
\caption{Method overview. The figure illustrates the architecture of the proposed OnlineHOI. OnlineHOI consists of the Mamba block and the Memory Augment block. The Mamba Encoder and Decoder both have an unidirectional spatial Mamba block which possesses forward scans within SSM layers respectively. The Memory Augment Model is placed between the Encoder and Decoder to enhance the key knowledge from the Encoding state and then pass it forward to the Decoder.}
\label{fig:overview}
\end{figure*}

In this section, we delineate the pipeline for both generation and perception in the online setting. We propose two new tasks: online generation and online perception. For online generation, the goal is to generate human or hand motion conditioned on object geometry and other human or hand states. For perception, we take a 4D point cloud video as input and output action categories. We illustrate our approach in Figure \ref{fig:overview}. To address the challenges of long-term instability and the scarcity of online knowledge, we propose OnlineHOI-G and OnlineHOI-P, which respectively generate realistic 3D HOI motion sequences and perceived HOI actions. Following this, We detail our OnlineHOI framework in two principal components: the Mamba Model in Sections~\ref{preli} and~\ref{Mamba backbone}, which focuses on temporal and spatial states, and the Memory Augment Model in Section~\ref{Memory}, which enhances forward pass performance.

\label{sec:method}

\subsection{Preliminaries}
\label{preli}
Mamba is a novel sequence modeling architecture that builds upon State Space Models (SSMs) while incorporating structured memory and efficient recurrence. It leverages exceptional performance in handling long-sequence tasks while maintaining high computational efficiency. These models represent the transformation of a one-dimensional function or sequence, \(x_t\in\mathbb{R}\), into an output \(y_t\in\mathbb{R}\) by utilizing a hidden state \(h_t\in\mathbb{R}^{N}\). The evolution of the hidden state is governed by the matrix \(\mathbf{A}\in\mathbb{R}^{N \times N}\), while \(\mathbf{B}\in\mathbb{R}^{N \times 1}\) and \(\mathbf{C}\in\mathbb{R}^{1 \times N}\) serve as projection matrices that map inputs to the hidden state and the hidden state to the output, respectively. Mamba contains a timescale parameter \(\boldsymbol {\Delta}\) to transform the matrices \(\mathbf{A}\), \(\mathbf{B}\) to \(\mathbf{\overline A}\), \(\mathbf{\overline B}\). After discretizing \(\mathbf{\overline A}\), \(\mathbf{\overline B}\), the Mamba model describe a system's evolution over time using a hidden state that is updated at each step:
\begin{equation}
\begin{aligned}
&h_{t+1} = \mathbf{\overline A} h_t + \mathbf{\overline Bx} _t, \\
&y_t = \mathbf{C} h_t.
\end{aligned}
\end{equation}
 At last, the output \(y\) is calculated through global convolution with a structured convolution kernel \(\mathbf{\overline K}\).
\begin{equation}
\begin{aligned}
&\mathbf{\overline K} = (\mathbf{C \overline B}, \mathbf{C\overline {AB}}, ..., \mathbf{C\overline A^{M-1}\overline B}), \\
&y = x*\mathbf{\overline K}.
\end{aligned}
\end{equation}
As the time sequence becomes longer, traditional Transformer models require storing the historical representation of the entire sequence for global self-attention calculations. In contrast, Mamba scans along the flow of time. Mamba’s structured state-space evolution, parameterized by \(\mathbf A\), \(\mathbf B\) and \(\mathbf C\), allows it to incrementally update its hidden state without the need to recompute previous steps, operating as a mode of state iteration. This provides natural advantages for online settings, where new inputs continuously arrive. It is particularly well-suited for integrating historical information to make decisions in the current frame.\\

\subsection{Mamba-based Backbone for Online HOI}
\label{Mamba backbone} 
For Transformers dealing with Online HOI, attention should be defined using a causal mask \cite{vaswani2017attention}, ensuring that the current frame can only perform self-attention with itself and previous frames. Since Transformer-based models rely on the attention mechanism to obtain both local and global information, this padding constraint leads to a significant drop in performance when applied to online setting tasks. In contrast, the Mamba-based model’s interactive state update system \cite{gu2023mamba} is more suitable for such settings. Based on this observation, we introduce OnlineHOI-P and OnlineHOI-G for online perception and generation, respectively.

\subsubsection{Mamba-based diffusion Model for online Generation}

Diffusion probabilistic models represent a major breakthrough in motion generation by leveraging a learned T-step Markov process \(\{x_t\}_{t}^{T}\) to progressively transform a Gaussian noise distribution into the target data distribution \(p(x)\) \cite{dhariwal2021diffusion, ho2020denoising}. Since Mamba model can handle long sequences data excellently, we build a mamba-based denoising model with long skip connections on the motion data. The diffusion on motion sequence is modeled as a Markov nosing process:
\begin{equation}
q(x_t \vert x_{t-1}) = \mathcal{N}(\sqrt{\alpha_t}x_{t-1}, (1-\alpha_t) I).
\end{equation}
where the constant \({\alpha_t} \in (0,1)\) are hyper-parameters, the \(\{x_t\}_{t=0}^{T}\) denotes the noising sequence, and \(x_{t-1}\) representing the previous step for denoising at the \(t\)-step. The architecture of the Mamba denoiser is illustrated in Figure. \ref{fig:overview}. At its core, it follows a U-Net-like structure comprising an encoder and a decoder. Both the encoder and the decoder contain an unidirectional spatial Mamba block to optimize unidirectional information propagation, enabling seamless top-down processing. Each sub-forward SSM scan first applies a 1-D convolution to the input, followed by a linear projection to transform it into \(\mathbf{\overline A}\) and \(\mathbf{\overline B}\). After executing a sequence of SSM forward scans, \(\mathbf{\overline A}\) and \(\mathbf{\overline B}\) iteratively update and generate the hidden state for each sequence. Finally, all sequences are computed by performing matrix multiplication on the output with the encoder or decoder input. To apply Mamba in the online setting, the scanning mechanism emphasizes improvements in representation learning and temporal scanning. The scanning mode is designed to move forward along the time dimension toward the future. At the end of the encoder, we build a self-attention block to condition and integrate the hidden state from the encoder output, enhancing conditional integration. Finally, the Mamba decoder decodes the hidden state and restores the motion sequence from Gaussian noise. We directly predict motion from Gaussian noise since our conditions are more complex than typical tasks, such as text or action conditions, which only involve a single low-dimensional input. We need to integrate both human or hand and object motion sequences into our conditional Mamba model. 
\subsubsection{Mamba-based 4D Model for online Perception}
We denote a point cloud sequence as \(\Psi = \{(P_t, V_t) \mid t = 1, \dots, L\}\), where \(P_t\) represents the point cloud at time step \(t\), and \(V_t\) denotes the associated normal vectors. 
Our 4D backbone adopts a UNet-like architecture. In line with the state-of-the-art P4Transformer \cite{fan2021point}, both the encoder and decoder consist of four layers of 4D convolution and deconvolution operations. For a given input clip $\Psi$, the convolution operation can be formally expressed as:\\
\begin{equation}
\begin{split}
\boldsymbol{f}_{t}^{\prime(x, y, z)}
 & = \sum_{\delta_{t}=-r_{t}}^{r_{t}} \sum_{\left\|\left(\delta_{x}, \delta_{y}, \delta_{z}\right)\right\| \leq r_{s}} \\ &
 (\boldsymbol{W_d}\cdot\left(\delta_{x}, \delta_{y}, \delta_{z}, \delta_{t}\right)^T) \odot 
(\boldsymbol{W_f} \cdot \boldsymbol{f}_{t+\delta_{t}}^{\left(x+\delta_{x}, y+\delta_{y}, z+\delta_{z}\right)})
\end{split}
\end{equation}
where \((x, y, z) \in P_t\) denote a point in the spatial domain at time \(t\), and \((\delta_x, \delta_y, \delta_z, \delta_t)\) represent the spatial-temporal offset of the convolution kernel. The operator \(\cdot\) denotes matrix multiplication. The feature at location \((x, y, z, t)\) is given by \(f_t^{(x, y, z)} \in \mathbb{R}^{C \times 1}\). Temporal aggregation is performed via sum-pooling across the temporal neighborhood, while spatial aggregation is conducted using max-pooling. The parameters \(r_s\) and \(r_t\) define the spatial and temporal receptive field radii, respectively. The offset-based weighting is computed as \(\boldsymbol{W_d} \cdot (\delta_x, \delta_y, \delta_z, \delta_t)^T\), where \(\boldsymbol{W_d} \in \mathbb{R}^{C' \times 4}\) maps the 4D displacement vector from \(\mathbb{R}^{4 \times 1}\) to a latent space of dimension \(\mathbb{R}^{C' \times 1}\). A projection matrix \(\boldsymbol{W_f} \in \mathbb{R}^{C' \times C}\) is then applied to align features across channels. The operator \(\odot\) denotes element-wise summation across the aggregated feature representations. Then the Mamba based OnlineHOI-P enhances per-point features obtained from the 4D backbone. The Mamba block enable the forward scanning through the timestep. The sub-forward-SSM scan calculates the \(\mathbf{\overline A}\), \(\mathbf{\overline B}\) from every sequence's point feature. Unidirectional scanning system iteratively updates every current state as the same as the generation setting.

\subsection{Memory Augment Model}
\label{Memory}

 Although the Mamba model performs significantly better on long sequence data, online settings for both perception and generation do not allow future information to be known in advance. This limitation can lead to inaccurate predictions, as relying solely on historical knowledge to forecast future data may be insufficient. To address this, we propose a Memory Augment Model to enhance the model's ability to predict the future in online setting. The Memory Augment Model produces \(\mathbf M_S\) (short-term memory) and \(\mathbf M_L\) (long-term memory) blocks. \\
\textbf{\(\mathbf M_S\) Memory}. In the online generation setting, \(\mathbf M_S\) stores the current time \(t\) and the previous frames. We set a fixed length of \(S\) frames to maintain this system which needs to be updated when the current time \(t>S\). Once the capacity is full, we build a update strategy based on the First-in-First-out (FIFO) queue. The earliest frame is cleared, and the current frame is stored, acting as a sliding window that flows with time. For initialization, when \(t<S\), we copy the frame at time \(t\) to the buffer until the capacity reaches \(S\). \(\mathbf M_S\) memory focuses on the previous information, which includes the current time \(t\) and its nearest \(S\) frames.\\
\textbf{\(\mathbf M_L\) Memory}. As time progresses, we introduce \(\mathbf M_L\) memory to reinforce earlier memories that are connected to the present moment, mitigating the problem of knowledge forgetting. \(\mathbf M_L\) memory first inherits the \(\mathbf M_S\) memory and then performs a series of filtering strategy operations. As shown in Algorithm \ref{algorithm1}, we calculate the dot product similarity between every two adjacent frames and store them in the similarity matrix, \(Sim\), to screen for more representative forward memories. Our objective is to preserve \(L\) frames (\(L\) is the \(\mathbf M_L\) capacity) after each merge operation, thereby retaining the rich information encapsulated within the \(\mathbf M_L\) memory. Accordingly, we iteratively merge adjacent frame pairs exhibiting the highest similarity through a weighted averaging process until the memory capacity reaches \(L\).
\renewcommand{\algorithmicrequire}{\textbf{Input:}}
\renewcommand{\algorithmicensure}{\textbf{Output:}}
\begin{algorithm} 
	\caption{Storage \(\mathbf M_L\) Memory}     
	 \label{algorithm1}       
	\begin{algorithmic}[1] 
	\Require \(\mathbf M_S\) memory 
    \Ensure  \(\mathbf M_L\) memory   
    \While {\(len(\mathbf M_S < L)\)}
    \State \(Sim = []\)    
    \For {\(t\) in \(S-1\)} 
        \State  Sim.append(dot(\(frame_t, frame_{t+1}\)))\
   \EndFor
   \State  \((t_{max}) = Max_{index}(Sim)\)
   \State  \(frame_{t_{max}} = (frame_{t_{max}}+frame_{{t_{max}}+1})/2\)
   \State del \(frame_{{t_{max}}+1}\)
   \EndWhile
   \end{algorithmic} 
\end{algorithm}\\ 
\textbf{\(\mathbf M_E\) Memory}. Note that we design exclusive \(\mathbf M_S\) memory and \(\mathbf M_L\) memory for each moment. Finally, we integrate these two different knowledge types by concatenating them to form the enhanced memory \(\mathbf M_E=[\mathbf M_S~ \vert~ \mathbf M_L]\). We encode the memory information for the output hidden state from the Mamba Encoder to obtain the enhanced memories \(\mathbf M_E\). The dimension of \(\mathbf M_E\) increases by one level because we calculate both \(\mathbf M_S\) and \(\mathbf M_L\) for each moment. Therefore, We utilize a Max-Pooling operation to map the dimension to the hidden state. Finally, we concatenate the \(\mathbf M_E\) and hidden state to form the input for the Mamba Decoder.

\section{Experiment}
\label{sec:exp}

We evaluate our method on both perception and generation for online HOI. We propose two novel tasks and validate them experimentally. The first task is the online HOI generation task, in which the objective is to generate the reactor’s actions based on the actor’s current and historical actions. Specifically, in the CORE4D dataset, the task involves online generation of full-body actions for the reactor from the actor’s full-body actions, whereas in the OAKINK2 dataset, it involves online generation of hand actions for the reactor from the actor’s hand actions.  We assume that the object pose is provided together with the actor’s state. This setup aligns with real-world scenarios, where the reactor needs to decide how to interact with the actor in an online setting. In the HOI4D perception task, the goal is to predict, in real time, the category of hand motion interactions at the current moment based on both current and historical inputs. In this Section, we first introduce the dataset settings, evaluation metrics, and implementation details. Then, we conduct both quantitative and qualitative evaluations to compare our method with others. Finally, we perform an ablation study to demonstrate the effectiveness of each module.

\subsection{Datasets}
We evaluate OnlineHOI on two prominent multi-human or objects interaction datasets and one contains 3D annotations of HOI dataset as follows:\\ 
\textbf{CORE4D} \cite{liu2024core4d}, a recent dataset, focuses on collaborative human-object-human interaction, which contains 875 real-word motion sequences spanning six object categories. We divide these sequences into a training set and two distinct testing sets S1 and S2, where the S1 represents the seen object geometries and S2 corresponds to the unseen geometries which can validate the generalization of the model.\\
\textbf{OAKINK2} \cite{yang2022oakink} dataset focuses on bimanual object manipulation tasks for complex daily activities. It contains two hands and multi-objects (one to three) interaction. We use the official split with one training sets and testing sets. \\
 \textbf{HOI4D} \cite{liu2022hoi4d} dataset is a large-scale, ego-centric hand-object interaction dataset, offering comprehensive multi-modal annotations. It includes 3,863 point cloud sequences from 9 participants interacting with 800 distinct object instances across 16 categories within 610 indoor environments, with each sequence containing 150 frames. For action segmentation, we use the official split of HOI4D, comprising 2,971 training scenes and 892 test scenes, with frame-level action labels for 19 unique classes.

\subsection{Evaluation Metrics}
Throughout our experiments, we adopt standard evaluation metrics across multiple aspects. Generation Quality is assessed using the Fréchet Inception Distance (FID), which measures the difference in feature distributions of generated and ground-truth motions. And we use the diversity (DIV) metric to measure motion diversity, which calculates variance in features extracted from the motions. Recognition Accuracy (RA) denotes the action recognition accuracy of the model on synthesized motions which can evaluate high-level motion semantics and its distributions. For perception, we using the framewise accuracy (Acc) first to directly denote the action accuracy. Segmental edit (Edit) ditance denotes the difference between two frames. And a seires of F1 scores at the overlapping thresholds of 10\(\%\) (F1@10), 25\(\%\) (F1@25) and 50\(\%\) (F1@50) are utilized to denote the performance on HOI prediction.

\subsection{Comparative Studies}
\subsubsection{Results on Generation Tasks}
\begin{figure*}
\centering
\includegraphics[width=0.9\textwidth]{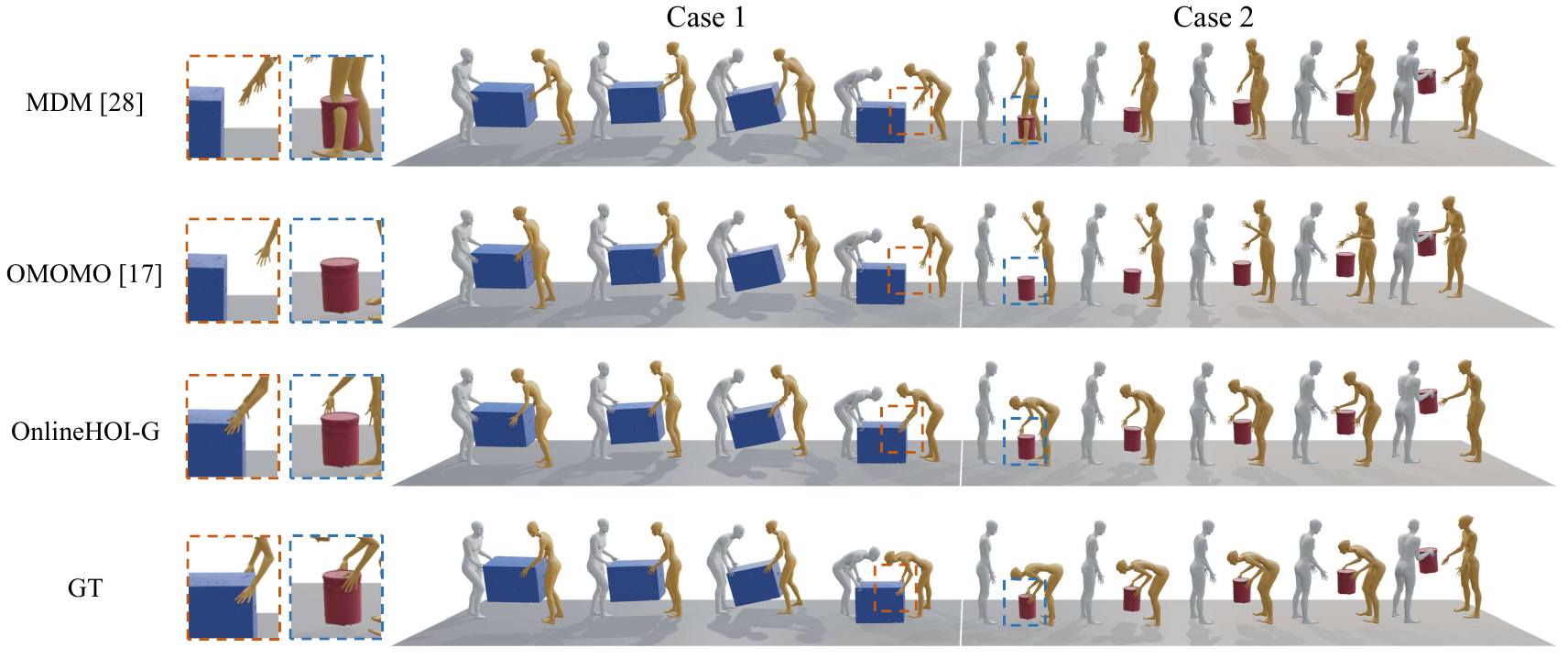}
\caption{Qualitative comparisons on CORE4D dataset. The yellow person represent the reactor which is the generation target, while the actor and object serve as conditions for the diffusion model. In Case 1, the reactors generated by MDM and OMOMO show almost no interaction with the object, while OnlineHOI-G is able to generate scenes where the reactor interacts with the object and cooperates with the actor. In Case 2, the scene in MDM and OMOMO gives the impression that the object is flying on its own, whereas in OnlineHOI-G, it appears normal for the reactor to pick up the object.}
\label{fig:CORE4D}
\vspace{-4mm}
\end{figure*}

\begin{figure*}
\centering
\includegraphics[width=0.9\textwidth]{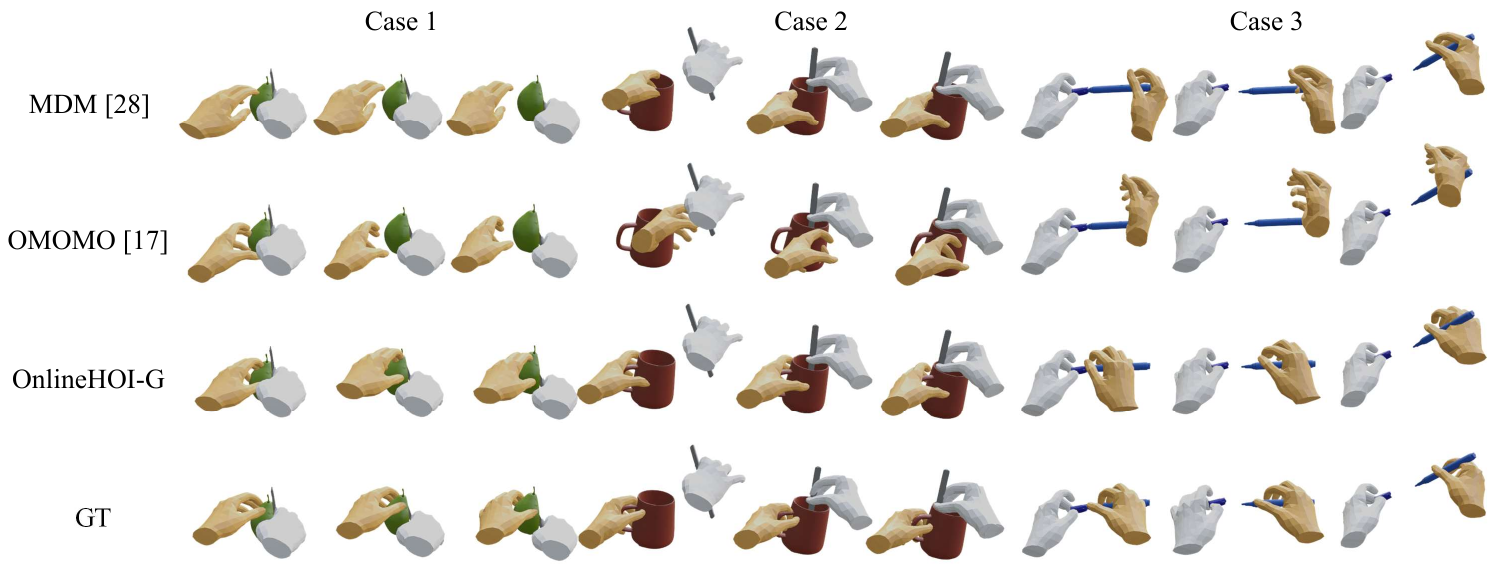}
\caption{Qualitative comparisons on OAKINK2 dataset. The yellow hand represents the reactor, which is generated by MDM, OMOMO, and OnlineHOI-G. In Case 1, the reactor generated by MDM and OMOMO can not make contact with the object, while the reactor generated by OnlineHOI-G maintains a relatively tight grip. In Case 2, MDM and OMOMO cause some object penetration, with OMOMO being more pronounced in this regard, whereas OnlineHOI-G does not exhibit this issue. In Case 3, both MDM and OMOMO fail to hold the pen properly, while OnlineHOI-G successfully handles this task.}
\label{fig:OAKINK2}
\vspace{-4mm}
\end{figure*}

We evaluate transformer-based methods, including MDM \cite{tevet2022human} and OMOMO \cite{li2023object}, alongside our method, OnlineHOI-G, on the CORE4D and OAKINK2 datasets for online generation. As shown in Tables \ref{tab:CORE4D} and \ref{tab:OAKINK2}, OnlineHOI-G outperforms MDM and OMOMO in terms of FID, RA, and DIV. The proportion of all User Study results exceeding half demonstrates the superiority of our method. Additionally, we conduct a series of visualizations to compare the generation quality of our method with MDM and OMOMO. As observed, OnlineHOI-G generates superior motion compared to the transformer-based methods. Figure \ref{fig:CORE4D} displays the multi-human-object interaction results, and multi-hand and object interactions are shown in Figure \ref{fig:OAKINK2}. Thus, our refined OnlineHOI-G is better suited for online settings. Both quantitative and qualitative results show that our method outperforms all others, proving that, by integrating our additional memory-augment model, OnlineHOI-G achieves remarkable online generation performance. OnlineHOI-G effectively enhances historical knowledge and extracts the most relevant information, enabling the Mamba model to perform better at the current moment.\\   
\vspace{-3mm}
\begin{table}[h!]
    
    \centering
    \resizebox{\columnwidth}{!}{
    \begin{tabular}{c|c|cccc}
    \hline
        {Test Set} & Method  & FID\(\downarrow\) & RA\(\uparrow\) & DIV\(\rightarrow\) 
        & User Study \(\%\) \\
        \hline
            \multirow{4}{*}{S1}
             & GT                         & 0.00 & 96.70 & 7.78 & -\\
             & MDM \cite{tevet2022human}                  & 3.56 & 77.77 & 7.87 & 19.79\\
             & OMOMO \cite{li2023object}               & 2.72 & 76.59 & 7.93 & 16.67\\
             & OnlineHOI-G          & \textbf{2.61} & \textbf{79.52} & \textbf{7.75} & \textbf{63.54}\\
        \hline
            \multirow{4}{*}{S2}
             & GT                         & 0.00 & 96.04 & 15.74 & - \\
             & MDM \cite{tevet2022human}                  & 3.73 & 76.56 & 15.86 & 21.43 \\
             & OMOMO \cite{li2023object}                & 3.11 & 75.86 & 15.90 & 26.79 \\
             & OnlineHOI-G          & \textbf{2.54} & \textbf{79.16} & \textbf{15.64} & \textbf{51.78} \\
        \hline    
    \end{tabular}}
    \caption{Results on Generation evaluated on CORE4D dataset. S1 and S2 denote seen and unseen objects, respectively. All models are trained and evaluated on the CORE4D dataset under identical settings, conditioned on both actor motion and object geometry. The reported metric evaluates the quality of generated reactor motion, which is the primary objective of these models.}
    \vspace{-6mm}
    \label{tab:CORE4D}
\end{table}
\vspace{-4mm}
\begin{table}[h!]
    \centering
    \resizebox{\columnwidth}{!}{
    \begin{tabular}{c|c|ccccc}
    \hline
        {Method}  & Clips/s & Acc & Edit & F1@10 & F1@25 & F1@50 \\
        \hline
        PointNet++ \cite{qi2017pointnet++} & \textbf{35.8} & 50.8  & 40.2 & 42.6 & 35.3 &23.5  \\
        \hline
        {P4Transformer \cite{fan2021point}} & 26.6 & 66.7 & 62.0  & 65.3  & 59.8 & 46.3 \\
        {OnlineHOI-P} & {31.8} & \textbf{71.2}  & \textbf{79.3} & \textbf{78.6} & \textbf{74.1} & \textbf{62.2} \\
    \hline
    \end{tabular}}
    \caption{Results on Action Segmentation Perception evaluated on HOI4D dataset. PointNet++ represents the direct use of single-frame information for prediction, while the P4Transformer/OnlineHOI-P incorporate different temporal layers on top of the single-frame features from PointNet++.}
    \vspace{-4mm}
    \label{tab:Pilot}
\end{table}

\begin{table}[th]
    
    \centering
    
    \begin{tabular}{cc|ccc}
    \hline
         & Method  & FID\(\downarrow\) & DIV\(\rightarrow\) & User Study \(\%\) \\
        \hline
             & GT                   & 0.02  & 15.59  & -\\
             & MDM \cite{tevet2022human}                  & 0.41  & 15.38 & 27.06\\
             & OMOMO \cite{li2023object}               & 0.43  & 15.46 & 18.82\\
             & OnlineHOI-G    & \textbf{0.35}  & \textbf{15.52} & \textbf{54.12}\\
        \hline    
    \end{tabular}
    \caption{Results on Generation evaluated on OAKINK2 dataset. We do the official train, validation and test split for these three models. The condition is uniformly consists of the actor's motion and two objects geometry. We evaluate the reactor's generation errors.}
    \vspace{-8mm}
    \label{tab:OAKINK2}
\end{table}

\begin{table*}[h!]
    \centering
    \resizebox{0.95\textwidth}{!}{
    \begin{tabular}{c|c|ccccc!{\vrule width 1pt}c|cc}
    \hline
        {Offline-Perception} & Clips/s & Acc & Edit & F1@10 & F1@25 & F1@50 &{Offline-Generation} & FID & RA  \\
        \hline
        
        {PointNet++ \cite{qi2017pointnet++}} & \textbf{35.8} & 50.8  & 40.2 & 42.6 & 35.3 &23.5 
        & {MDM (Transformer) \cite{tevet2022human}} & {\textbf{2.89}} & {\textbf{80.36}}  \\

        \hline
        {Transformer \cite{fan2021point}} & 7.0 & \textbf{71.2}  & \textbf{73.1} & \textbf{73.8} & \textbf{69.2} & \textbf{58.2}
        & \multirow{2}{*}{Mamba} & \multirow{2}{*}{3.05} & \multirow{2}{*}{79.01} \\
        {Mamba} & {17.7} & 66.0 & 65.4 &69.0  & 63.3 & 50.8 &\multirow{2}{*}{}\\
        \noalign{\hrule height 1pt}    
        {Online-perception} & Clips/s & Acc & Edit & F1@10 & F1@25 & F1@50 & {Online-Generation} & FID & RA \\
        \hline
        {PointNet++ \cite{qi2017pointnet++}} & \textbf{35.8} & 50.8  & 40.2 & 42.6 & 35.3 &23.5
        & {MDM (Transformer) \cite{tevet2022human}} & {3.73} & {76.56} \\
        \hline
        {Transformer \cite{fan2021point}} & 26.6 & 66.7 & 62.0  & 65.3  & 59.8 & 46.3 
        & \multirow{2}{*}{Mamba} & \multirow{2}{*}{\textbf{2.72}} & \multirow{2}{*}{\textbf{78.26}} \\
        {Mamba} & {33.4} & \textbf{71.0}  & \textbf{78.5} & \textbf{77.8} & \textbf{73.4} & \textbf{61.5} & \multirow{2}{*}{} \\


    \hline
    \end{tabular}
     }
    \caption{Online and offline results on Transformer and Mamba based Model.}
    \vspace{-8mm}
    \label{tab:Online and Offline}
\end{table*}

\subsubsection{Results on Perception Tasks}
The experimental results highlight the substantial advantages of OnlineHOI-P over previous methods. Comparing OnlineHOI-P with PointNet++, we find that adding a lightweight Mamba temporal fusion layer doubles performance, underscoring the value of transitioning from PointNet++ to OnlineHOI-P in online perception. While comparing on P4Mamba, the Memory Augment Model take the Mamba model's performance to the next level. The OnlineHOI-P approach significantly outperforms existing methods and achieves superior results compared to  PointNet++ and P4Transformer. This result clearly demonstrates the efficiency and effectiveness of OnlineHOI-P. The significant improvement of OnlineHOI-P in Edit and F1 scores indicates that it has also enhanced the smoothness and consistency of its network predictions.

\subsection{Perceptual User Study}
We conducted a perceptual user study with approximately 20 participants to evaluate the human and hand motion sequences generated by OnlineHOI-G, MDM, and OMOMO. The user study included a diverse set of 40 motion sequences, randomly selected from the Core4D and OAKINK2 test sets. We calculate the proportion of votes for each method out of the total. The results are presented in Tables \ref{tab:CORE4D} and \ref{tab:OAKINK2}. OnlineHOI-G received more than half of the votes, indicating that our method generates superior quality in the online setting.

\subsection{Ablation studies}
We execute the offline and online experiment respectively on HOI generation and perception. Then we examine the key deign Memory Augment Model for different setting.\\ 
\vspace{-4mm}
\subsubsection{Online and Offline}
We compare the Online and Offline on perception and generation. We execute transformer-based and mamba-based model to find which part they are adept at. The Table \ref{tab:Online and Offline} shows that transformer-based model is more suitable for offline perception and generation. While on online tasks, mamba-based model has greater advantages over transformer-based model. Transformer-based model's performance drop obviously when handle the online task because self-attention's ability will be constraint since online setting can not do global self-attention across the whole timeline.  
For Mamba-based model, since Mamba modeling with long sequences make it the stability of global time series performs better than Transformer. Mamba's better skill on handling long sequence motion on online setting make the generation quality better than transformer based model.

\subsubsection{w/ or w/o Memory}
We evaluate the performance of our method with and without the Memory Augment Model. As shown in Tables \ref{tab:Results on w/ and w/o Memory on generation.}, \ref{tab:Results on w/ and w/o Memory on perception.}, the OnlineHOI model, which includes the \(M_E\) Memory, performs better than the OnlineHOI model without it. The Memory Augment Model we designed provides more critical knowledge, which enhances the stability of the Mamba model's scanning and optimizes its inference for future scenarios. Thus, OnlineHOI can leverage its full potential to handle both generation and perception tasks in an online setting. \\

\begin{table}
    \centering
    
    \begin{tabular}{c|ccc}
    \hline
         {Method} & FID\(\downarrow\) & RA\(\uparrow\) & {DIV (GT~15.74)} \\
        \hline
              OnlineHOI-G (w/o)  & 2.72  & 78.26 & 15.51  \\ 
              OnlineHOI-G (ours)   & \textbf{2.54}  & \textbf{79.16} & \textbf{15.64} \\
        \hline   
    \end{tabular}
    \caption{Results on w/ and w/o Memory on generation.}
   \vspace{-8mm}
    \label{tab:Results on w/ and w/o Memory on generation.}
\end{table}

\begin{table}
    \centering
    
    \begin{tabular}{c|ccccc}
    \hline
         {Method} & {Acc} & {Edit} & {F1@10} & {F1@25} & {F1@50} \\
        \hline
              OnlineHOI-P (w/o)  & {70.5}  & {78.5} & {77.8} & {73.4} & {61.5}  \\ 
              OnlineHOI-P (ours)   & \textbf{71.2} & \textbf{79.3} & \textbf{78.6} & \textbf{74.1} & \textbf{62.2}\\
        \hline   
    \end{tabular}
    \caption{Results on w/ and w/o Memory on perception.}
   \vspace{-8mm}
    \label{tab:Results on w/ and w/o Memory on perception.}
\end{table}

\vspace{-4mm}
\subsubsection {\(\mathbf M_S\) Memory, \(\mathbf M_L\) Memory or \(\mathbf M_E\) Memory and Memory Fusion}
We compare with three different Memory setting, only with\(\mathbf M_S\) Memory, only with\(\mathbf M_L\) Memory and \(\mathbf M_E\) Memory respectively on CORE4D dataset. And we design three methods in add fusion, max fusion and concatenation with max pooling which is our method's setting. As the Table \ref{tab:CORE4D memory ablation} shown, both the \(\mathbf M_S\) and \(\mathbf M_L\) Memory alone could not reach the full version of our Memory Enhanced Model, i.e. \(\mathbf M_E\) Memory. The \(\mathbf M_E\) Memory consists of \(\mathbf M_S\) and \(\mathbf M_L\) Memory is the best Memory representation of history to make up for the deficiency of the Mamba model for online generated knowledge sources. And our fusion method can better integrate \(\mathbf M_E\) Memory with hidden state to facilitate Mamba Decoder to decode the motion information. \\

\vspace{-6mm}
\begin{table}[h!]
    \centering
    \resizebox{0.8\columnwidth}{!}{
   \begin{tabular}{c|c|cc!{\vrule width 1pt}c|cc}
    \hline
    \multicolumn{4}{c!{\vrule width 1pt}}{\textbf{\(\mathbf M_S\), \(\mathbf M_L\) and \(\mathbf M_E\) Memory}} &  \multicolumn{3}{c}{\textbf{Memory fusion}} \\
    \hline
        {Test Set} & Method & FID\(\downarrow\) & RA\(\uparrow\) & Method & FID\(\downarrow\) & RA\(\uparrow\) \\
        \hline
            & \(M_S\) & 2.76  & 78.56 & Add Fusion & 3.08  & 78.31\\
        {S1} & \(M_L\) & 2.64  & 79.15 & Max Fusion & 3.15  & 77.66\\
             & \(M_E\) & \textbf{2.61}  & \textbf{79.52} & Ours  & \textbf{2.61}  & \textbf{79.52}\\
        \hline
             & \(M_S\) & 2.84   & 78.48 & Add Fusion & 3.26  & 76.01\\
        {S2} & \(M_L\) & 2.75   & 77.66 & Max Fusion & 3.25  & 76.25\\
             & \(M_E\) & \textbf{2.54}   & \textbf{79.16} & Ours  & \textbf{2.54}  & \textbf{79.16}\\
        \hline    
    \end{tabular}}
    \caption{Results on \(\mathbf M_S\), \(\mathbf M_L\) , \(\mathbf M_E\) Memory and Memory fusion. S1 and S2 denote seen and unseen objects, respectively.}
    \vspace{-8mm}
    \label{tab:CORE4D memory ablation}
\end{table}

\vspace{-2mm}

\section{Conclusion}
\label{sec:conclusion}
In this paper, we propose OnlineHOI, a novel method designed to address the challenges of online HOI perception and generation. Unlike conventional Transformer architectures, our approach employs a unidirectional scanning mechanism integrated with both short- and long-term memory modules, enabling efficient capture of sequential historical information. Experimental evaluations on the Core4D, OKINK2, and HOI4D datasets demonstrate significant enhancements in key performance metrics across both generative and perceptual tasks. 


\clearpage

\bibliographystyle{plain}
\bibliography{ref}

\end{document}